\documentclass[smallcondensed]{svjour3}     
\smartqed  
\usepackage[colorlinks]{hyperref}
\usepackage{verbatim}
\usepackage{float}
\usepackage{mathtools}
\usepackage{mathptmx}
\usepackage{fullpage}

\usepackage[utf8]{inputenc} 
\usepackage[T1]{fontenc}    
\usepackage{hyperref}       
\usepackage{url}            
\usepackage{booktabs}       
\usepackage{amsfonts}       
\usepackage{nicefrac}       
\usepackage{microtype}      
\usepackage{xcolor} 
\usepackage{multirow}        
\usepackage{subfigure}
\usepackage{graphics}
\usepackage{graphicx}
\usepackage{amsmath}

\journalname{}

\makeatletter
\g@addto@macro{\UrlBreaks}{\UrlOrds}
\makeatother

\begin{document}

\title{Concordance based Survival Cobra with regression type weak learners} 

\titlerunning{ConSurvCoBRA}        
\author{ 
Rahul Goswami and Arabin Kumar Dey }

\institute{Rahul Goswami \at
             Department of Mathematics,\\ 
             IIT Guwahati,\\   
             Guwahati, India\\
             \email{rahul.goswami@iitg.ac.in} \\
             \and
             Arabin Kumar Dey \at
             Department of Mathematics,\\ 
             IIT Guwahati,\\   
             Guwahati, India\\
             \email{arabin@iitg.ac.in} \\
}

\maketitle

\begin{abstract}

  In this paper, we predict conditional survival functions through a combined regression strategy.  We take weak learners as different random survival trees.  We propose to maximize concordance in the right-censored set up to find the optimal parameters.  We explore two approaches, a usual survival cobra and a novel weighted predictor based on the concordance index.  Our proposed formulations use two different norms, say, Max-norm and Frobenius norm, to find a proximity set of predictions from query points in the test dataset.  We illustrate our algorithms through three different real-life dataset implementations. 

\end{abstract}

\section{Introduction}
\label{intro}

  Combined regression strategy (COBRA) [\cite{biau2013cobra}, \cite{biau2016cobra}] is a powerful ensemble technique which has its own orientation.  Majority of the work available in prediction of conditional survival function so far uses Adaboost type of ensemble and its variations [\cite{bellot2018boosted}, \cite{solomatine2004adaboost}, \cite{djebbari2008ensemble}].  No work is available for predicting conditional survival function using Combined regression strategy (COBRA).  

  In this paper, we explore the COBRA ensemble setup by maximizing the concordance index.  We choose different survival tree-based models as weak learners.  The Concordance index (C-index) extends the usage for right-censored data too.  This approach is a combination of several pure regressor-type models.  We use two norms (Frobenius and max) to check similarities between observations and compare the results.  Existing COBRA implementation in regression setup uses a one-dimensional response variable.  In survival regression, the response variable is the survival function.  Therefore it is represented by a vector of values.  COBRA structure implementation changes mainly due to this functional structure of the response variable.   

 There is a wide variety of ensemble algorithms [\cite{dietterich2000ensemble}, \cite{giraud2014introduction}, \cite{shalev2014understanding}], with a crushing majority devoted to linear or convex combinations.   Boosting techniques have other several variants used in different context, like Gradient Boosting [\cite{friedman2002stochastic}], SurvivalBoost.R, SurvivalBoost.T [\cite{bellot2018boosted}], XGboost [\cite{chen2015xgboost}] which is similar but different than COBRA.  A non-linear way of combining estimators is available by [\cite{mojirsheibani1999combining}].  The adaboost [\cite{solomatine2004adaboost}] also uses threshold parameters.  However, Adaboost uses an exponential error function, whereas original COBRA for normal classification and regression builds its integration through square error loss.  In Survival COBRA, we use Concordance loss instead of square error loss.

 Since the weak learners are Survival Trees we compare COBRA with similar ensemble structure like Random survival forest.  We study the amount of improvement obtained by combining all the weak-learners.  The current study does not compare among the different existing variations of the adaboost with current proposed structure of COBRA, since adaboost orientation is totally different and does not show better result than Random Survival Forest when C-index is taken as metric [see \cite{bellot2018boosted}].    

  We organize the paper in the following way.  In section 2, we provide the conventional structure of COBRA.   We propose a Survival COBRA structure to predict conditional survival in section 3.  An illustration of the real dataset is available in section 4.  We conclude the paper in section 5.

\section{Usual Cobra}

  Combined regression strategy (COBRA) combines predictions from different weak learners to make the prediction.  
Suppose we have a data set of the following form : $(X_{i}, y_{i})\in R^{d + 1}; i = 1, \ldots, N$.  We divide the data set into training and test set, where test set observations are, $(X_{i}, y_{i})\in R^{d + 1}; i = 1, \ldots, l$.  Let's denote the M weak learners that COBRA uses are, say $r_{1}(\cdot), r_{2}(\cdot), r_{3}(\cdot), \ldots, r_{M}(\cdot) : R^{d} \rightarrow R.$  COBRA comes with an objective to find an ensembled strategy to combine all the regressor to predict a new observation with better accuracy.  It uses its predictor at particular instances ($x$) as a weighted sum of those test samples $Y_{i}$ where weights are calculated based on some test sample instances, which are in the neighborhood of predictions made by each predictor at the particular instances. 

  Mathematically, the COBRA strategy expresses a predictive estimator $T_{n}$, which we can write as 

$$T_{n}(r_{k}(\textbf{x})) = \sum_{i = 1}^{l} W_{n, i}(\textbf{x}) Y_{i}, ~~~~~~ x \in R^{d}$$  

  where the random weights $W_{n, i}(\textbf{x})$ take the form :
$$ W_{n, i}(\textbf{x}) = \frac{1_{\cap_{m = 1}^{M} {|r_{k, m}(x) - r_{k, m}(X_{i})| \leq \epsilon_{l}}}}{\sum_{j = 1}^{l} 1_{\cap_{m =1}^{M} {|r_{k, m}(x) - r_{k, m}(X_{i})| \leq \epsilon_{l}} }}.$$

  Here $\epsilon_{l}$ is a positive number and $\frac{0}{0} = 0$ (by convention).  The parameter $\epsilon_{l}$ plays the role of smoothing parameters.  In addition, we can provide an alternative way to express COBRA when all original observations have the same, equally valued opinion on the importance of the observation $X_{i}$ (within the range of $\epsilon_{l}$).  Generally, the corresponding $Y_{i}$ forms $T_{n}$ in such cases.  However the unanimity constraint may be relaxed by imposing, for example, a fixed fraction $\alpha \in \{ \frac{1}{M}, \frac{2}{M}, \cdots, 1 \}.$ of the machines agrees on the importance of $X_{i}$.  In that case, the weights take a more sophisticated form :
$$ W_{n, i}(x) = \frac{1_{\sum_{m = 1}^{M} 1_{|r_{k, m}(x) - r_{k, m}(X_{j})| \leq \epsilon_{l} } \geq M\alpha}}{\sum_{j = 1}^{l} 1_{\sum_{m = 1}^{M} 1_{|r_{k, m}(x) - r_{k, m}(X_{j})| \leq \epsilon_{l} } \geq M\alpha}}  $$  

  Instead of counting the number of observations similar to the query point and expressing the same in terms of an indicator function, researchers used kernel approximation of the above expression as a smooth representation so that we can choose the optimal direction of all parameters.  

\textbf{Key Implementation Issues in above Context :}   

 Since the algorithm requires extraction of similar observations depending on a particular choice of threshold, it may not always be capable of selecting a non-zero number of similar samples.  This situation will arise as the threshold value comes closer to zero or exactly equals to zero.  A solution to the problem is to increase the test set size. Alternatively, the algorithm suggests using a proportion ($\alpha$) of machine predictions instead of considering all the machine predictions. 
 
\section{Proposed Survival COBRA}  

 The usual general regression setup uses the response variable/output variable with only one dimension.  The conditional survival prediction predicts a function, and thus output variable/prediction space is always a higher dimension (since a set of points represents the function).  Therefore usual COBRA structure does not work in a conditional survival prediction setup.  We propose two Survival COBRAs in this context.  

\subsection{Vanilla Survival Cobra}

 We perform the experimental procedure in the following way.  In the first step, we focus on the calculation of the parameters.  In the second step, we find the curve estimates and benchmark scores based on the optimized parameters.  We divide the whole set as train and test sets and start optimizing the parameters based on the train set for an arbitrary split division.  The algorithm will further subdivide the whole dataset into two parts, (say, $D_{l}$ and $D_{k}$), where all the weak-learners training use $D_{k}$ only.  In this process test set would be a query set or validation set.   The prediction uses $D_{l}$.  However, a small original dataset would create $D_{l}$ much smaller due to such division, which could be a bottleneck for getting the desired optimal parameters.  Therefore, we choose a slightly different way to perform the split division.  First, we use the whole dataset to find the optimal parameters and then split the dataset into training and test sets to get the survival estimate at optimized COBRA parameters.  

  Our proposed algorithm uses weak learners as different survival trees.   We take eight weak-learners and follow up on the corresponding results.   Since each weak-learners are different regressions producing survival functions, the output dimension is high (depends on the number of time points considered).  While aggregating similar observations in the test set, we first choose a suitable norm to measure the similarity.  We take two different norms in this case : (1) Frobenius norm and (2) Sup norm, which calculates the maximum of all prediction distances.  In general, the choice of the value of the threshold parameter ($\epsilon_{l}$) and a proportion of a total number of predictors ($\alpha$) depend on the minimization of the cross-validated test error.  Our current exploration uses the concordance index for right-censored data.  In a different note, we propose to choose this $\epsilon_{l}$ and $\alpha$ by maximizing concordance index measured for right censored data.  We study the other benchmarks in a separate work.

\textbf{Concordance Index }  The C-index or concordance index $C(t)$ for two patients $i$ and $j$ can be given by the following expression :  $C(t) = P(\hat{S}_{i}(t) > \hat{S}_{j}(t) | \delta_{i} = 1, t \leq T_{j}, T_{i} > T_{j})$
where, $T_{i}, T_{j}$ are time to event data for patients $i$ and $j$ respectively and $\hat{S}_{i}(t)$, $\hat{S}_{j}(t)$ are the conditional survival functions for patient $i$ and patient $j$ respectively.  Alternatively, we can choose to use $ C(t) = P(\hat{r}_{i} < \hat{r}_{j} | \delta_{i} = 1, t \leq T_{j}, T_{i} > T_{j}) $ where, $\hat{r}_{i}$, $\hat{r}_{j}$ are the risk scores corresponding to the $i$th and $j$-th patient.  We define a risk score as total sum of cumulative hazard function calculated across all predecided time points. 

 If the data is censored we choose $\delta_{i} = 1$.  Since, we propose to maximize C-index to train Survival COBRA, We can call the algorithm as CindexSurvCobraBoosting.

 C-index calculation takes the following steps:
 \begin{enumerate}
 
 \item After forming all possible pair cases, omit those pairs whose shorter survival time is censored. Omit pairs i
and j if $T_i = T_j$ unless at least one is a death. Let Permissible denote the
total number of permissible pairs.
 
 \item For each permissible pair where $T_i \neq T_j$ , count 1 if the shorter survival
time has worse risk score as defined above; count 0.5 if predicted outcomes are
tied. For each permissible pair, where $ T_i = T_j$ and both are deaths, count 1
if predicted outcomes are tied; otherwise, count 0.5. For each permissible
pair where $T_i = T_j$ , but not both are deaths, count 1 if the death has
worse predicted outcome; otherwise, count 0.5. Let Concordance denote
the sum over all permissible pairs.
 
 \item he C-index, C, is defined by C = Concordance/Permissible.
 
 \end{enumerate}

  Concordance intuitively means that two samples were ordered correctly by the model. More specifically, two samples are concordant, if the one with a higher estimated risk score has a shorter actual survival time. 
  
\subsection{Proposed Algorithm- Weighted Survival Cobra}

  One of the problems in the above Vanilla COBRA is the approach does not allow evaluating the survival curve properly for a given patient if a sufficient number of observations from the test set comes in proximity to that patient observation.  We suggest an alternative procedure to find the survival curve rectifying the problem of an insufficient number of observations in the proximity of the query of the point.  The proposed methodology works as long as there is a non-zero observation available in the proximity of the query point. 
  
Key algorithmic steps of the weighted version COBRA are as follows :

\begin{enumerate}  

\item Find optimized parameters from the data set with five-fold cross-validation.

\item Divide the data set into train and test.   Use test dataset as query point and further divide train dataset into two subdivisions where the we train the weak learners on one such subdivision and prediction is made based on the other.

\item Collect all similar points in the proximity of the query point based on the chosen optimized parameters, (there should be at least one).

\item Use machine prediction (weak learners) as an average of all the predictions on the set of similar points.

\item Final prediction is the weighted average of all machine predictions where weights are taken as normalized concordance index by each machine. 

\end{enumerate}

 The above approach does not hurt the original notion of COBRA as it uses the contribution of the observed response variables in the weight calculation.  The final prediction ensures that the algorithm generates a survival curve or cumulative hazard function even when only one observation is available in the proximity setting.

\textbf{Key issues/Limitations with Weighted Survival Cobra}  Although Weighted Survival COBRA is capable of generating the survival curve in both the suggested norms, its performance depends on the choice of the optimal parameters.  The algorithm requires at least one observation in the proximity of the query point.  The computational run time increases with more inclusion of machines.               

\section{Data Analysis}

\subsection{Data Set Descriptions}  We have taken three datasets for experimentation 
\begin{itemize}
    \item Dataset I :  Worcester Heart Attack Study dataset \label{data2}, it has 500 samples and 14 features. Total number of deaths recorded are for 215 patients.
    \item Dataset II :  German Breast Cancer Study Group 2 dataset \label{data4} : The dataset has 686 samples and 8 features.
    The endpoint is recurrence free survival, which occurred for 299 patients (43.6\%).  Response variables are :
    $cens$: Boolean indicating whether the endpoint has been reached or the event time is right censored.    
    $time$: Total length of follow-up.
    \item Dataset III : Veteran Lung Cancer Data \label{data1}, it has 137 samples and 6 features out of which number of deaths recorded are 128.
\end{itemize}

 All datasets are publicly available and extracted directly from python module \textit{sksurv}.
      
\subsection{Numerical Results}  

  All codes are run at IIT Guwahati computers with the following configurations : (i) Intel(R) Core(TM) i5-6200U CPU 2.30 GHz.  (ii) Ubuntu 20.04 LTS OS, and (iii) 12 GB Memory.  All codes in written in python 3.9 in Jupyter Notebook.  All codes can be made available on request to authors.

\begin{figure}[ht!]
 \begin{center}
  \subfigure[$\xi_{1}$]{\includegraphics[width = 0.45\textwidth]{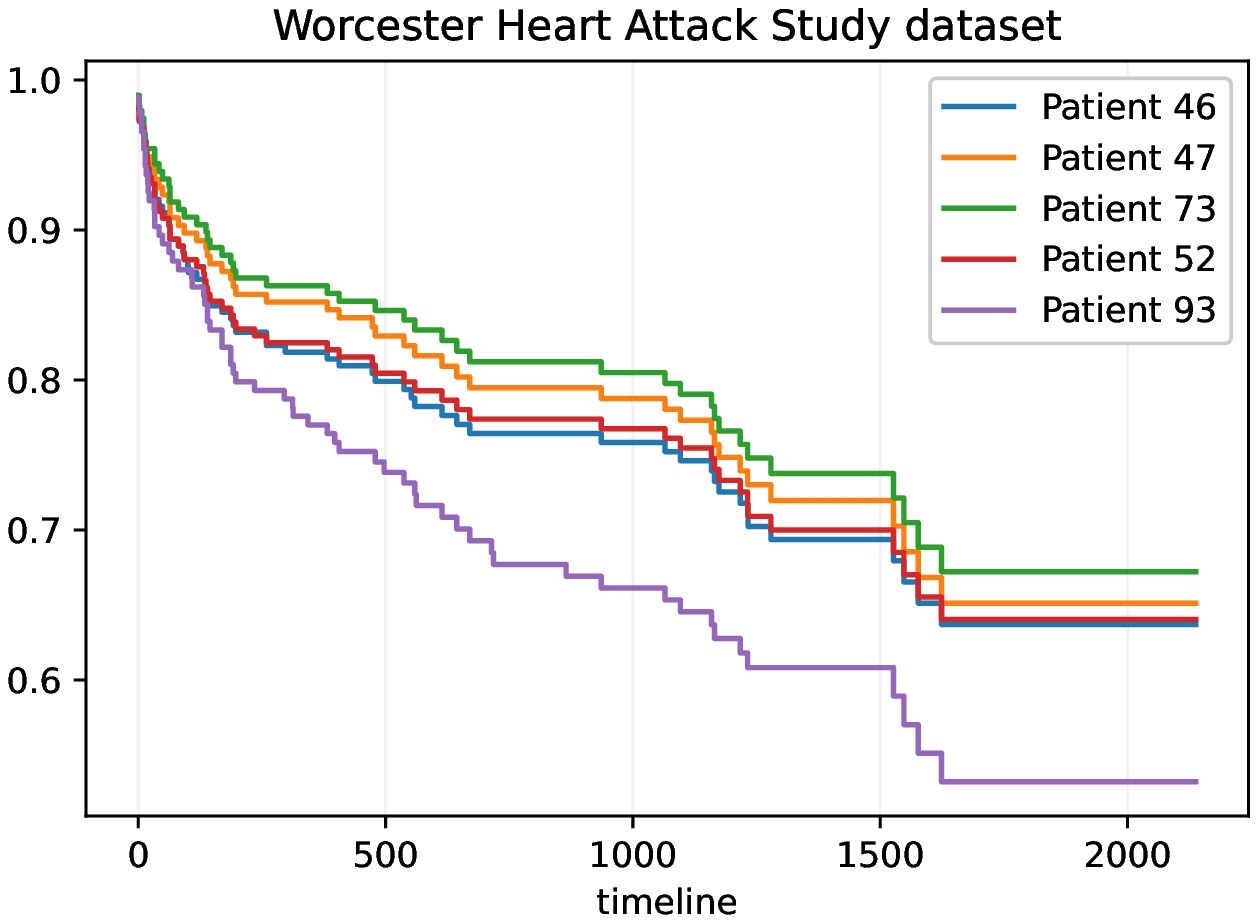}}
  \subfigure[$\xi_{2}$]{\includegraphics[width = 0.45\textwidth]{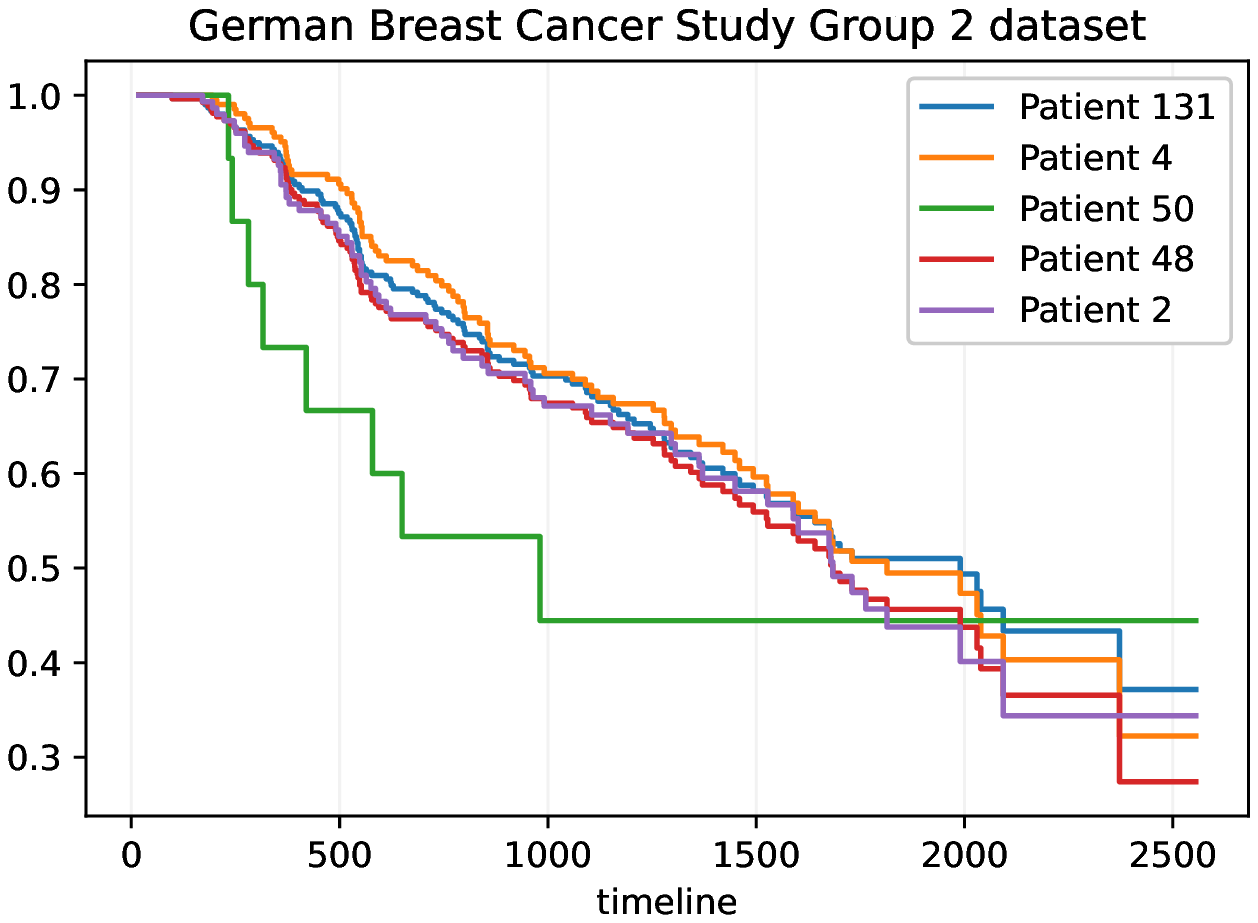}}\\
  \subfigure[$\xi_{3}$]{\includegraphics[width = 0.65\textwidth]{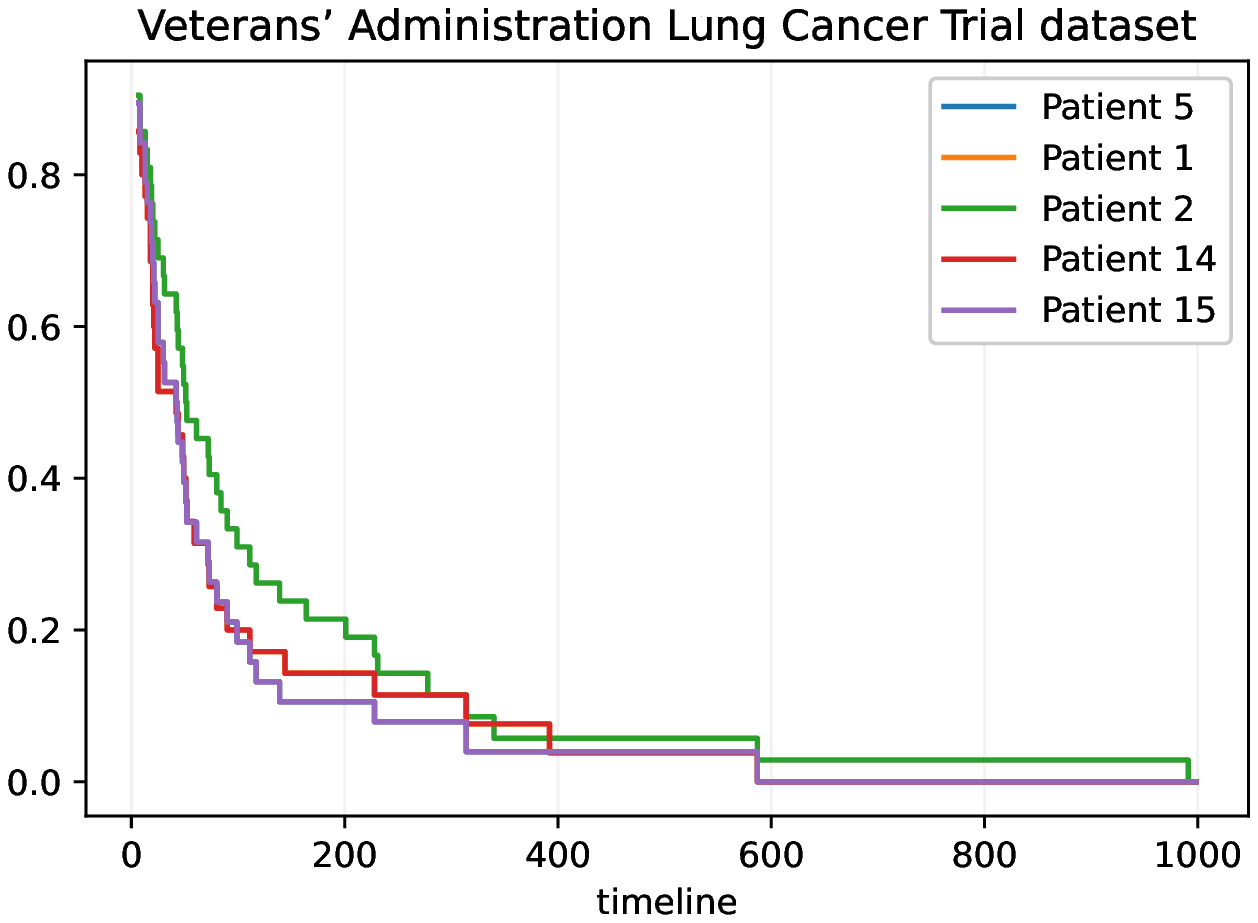}}\\
\caption{Survival Curve for Vanilla Survival Cobra for three datasets with Frobenius Norm \label{fig1}}
\end{center}
\end{figure}

\begin{figure}[ht!]
 \begin{center}
  \subfigure[$\xi_{1}$]{\includegraphics[width = 0.45\textwidth]{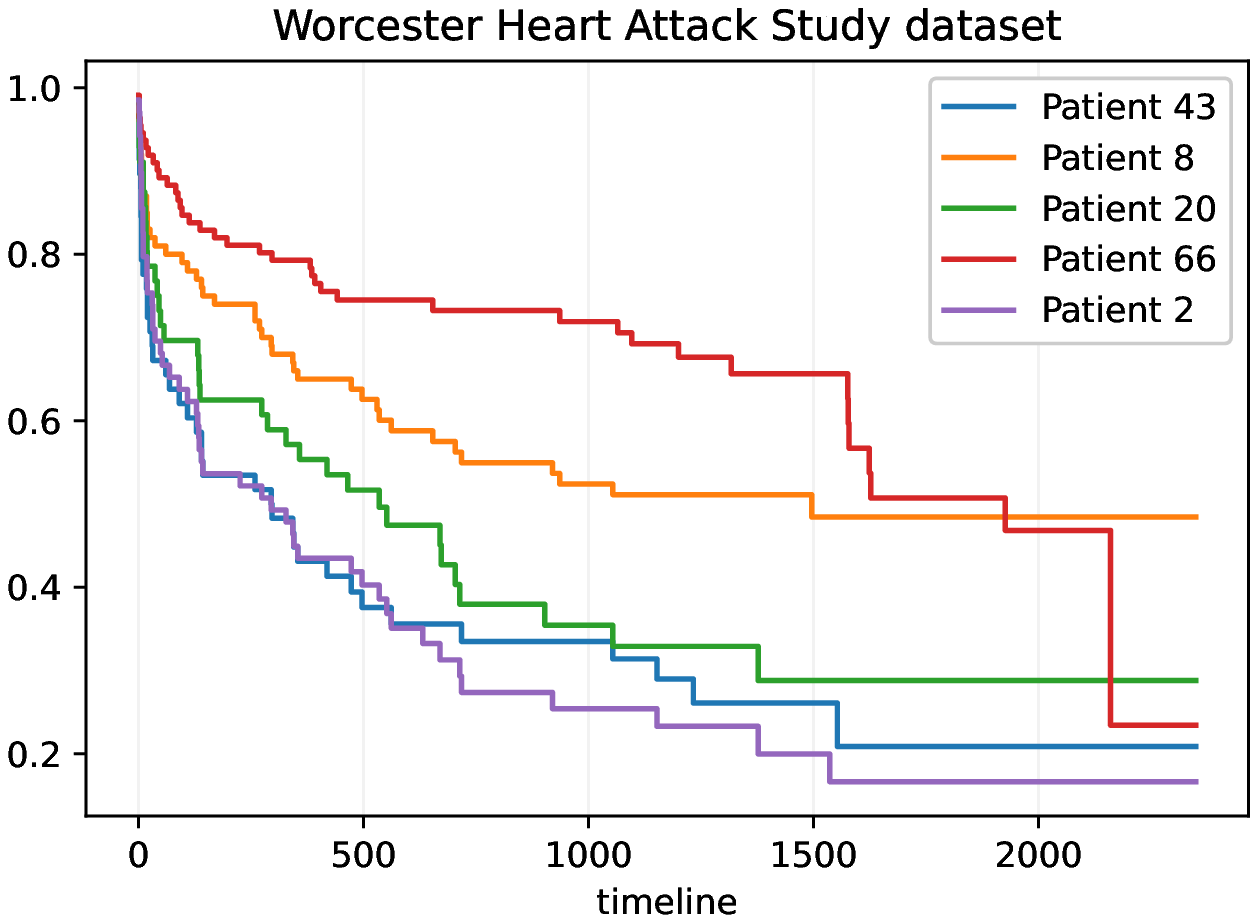}}
  \subfigure[$\xi_{2}$]{\includegraphics[width = 0.45\textwidth]{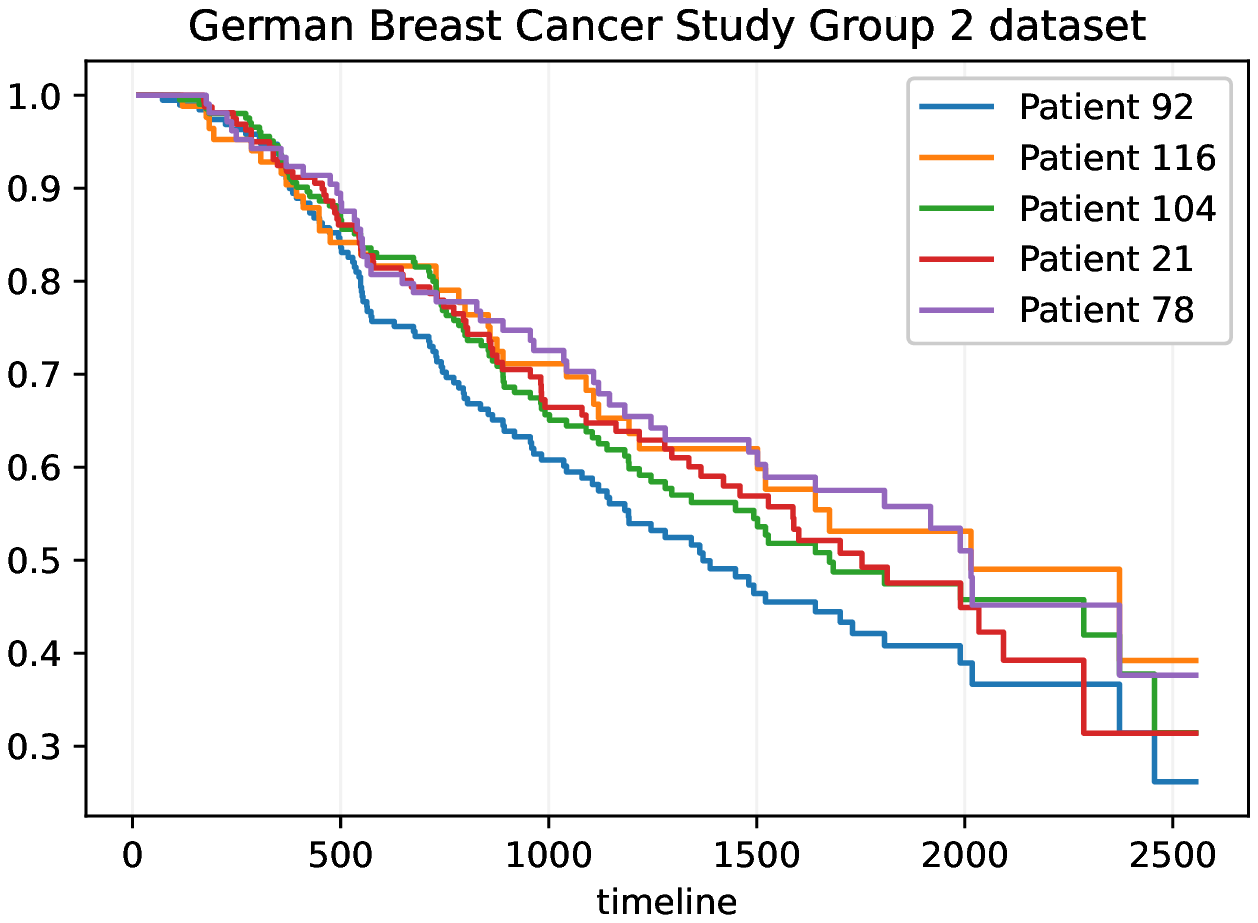}}\\
  \subfigure[$\xi_{3}$]{\includegraphics[width = 0.65\textwidth]{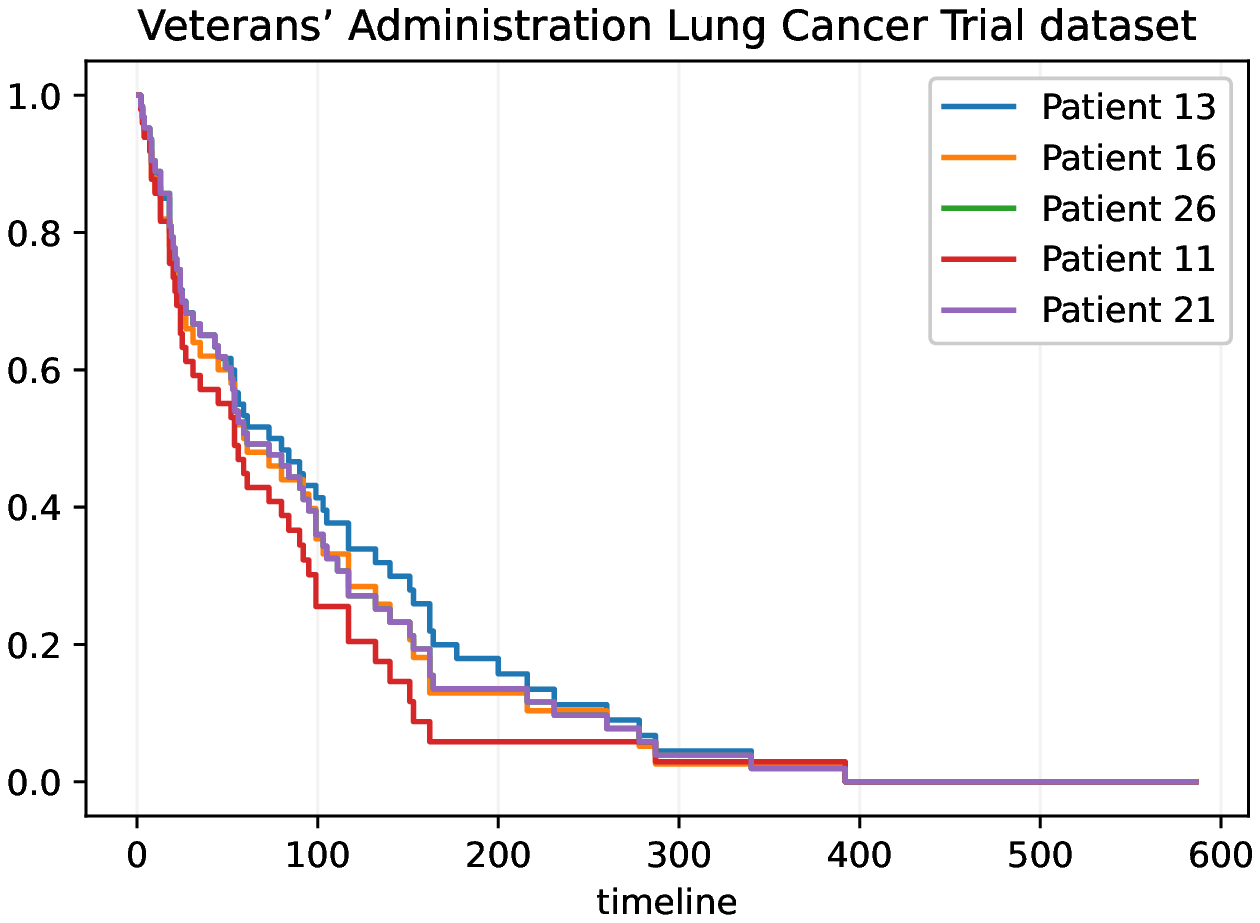}}\\
\caption{Survival Curve for Vanilla Survival Cobra for three datasets with Max Norm \label{fig2}}
\end{center}
\end{figure}

\begin{figure}[ht!]
 \begin{center}
  \subfigure[$\xi_{1}$]{\includegraphics[width = 0.45\textwidth]{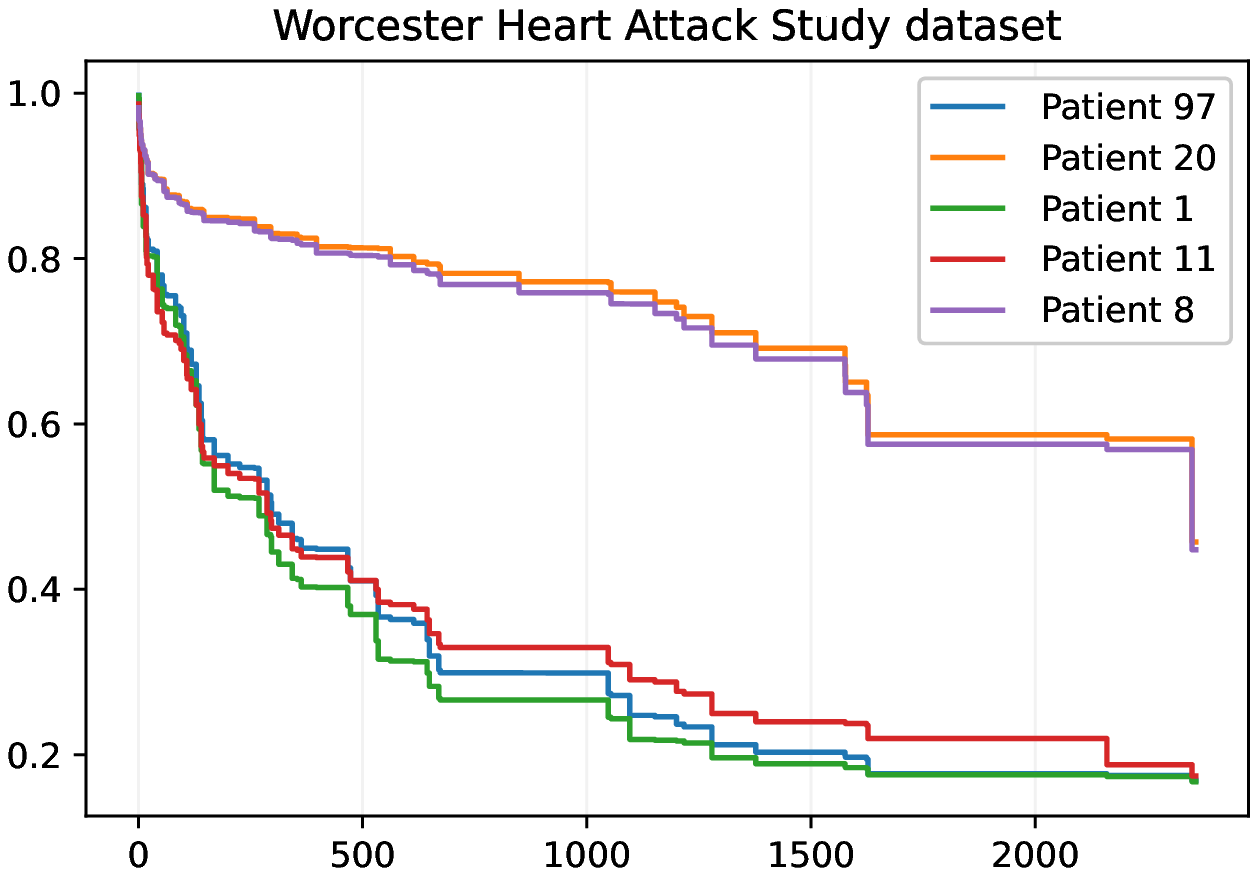}}
  \subfigure[$\xi_{2}$]{\includegraphics[width = 0.45\textwidth]{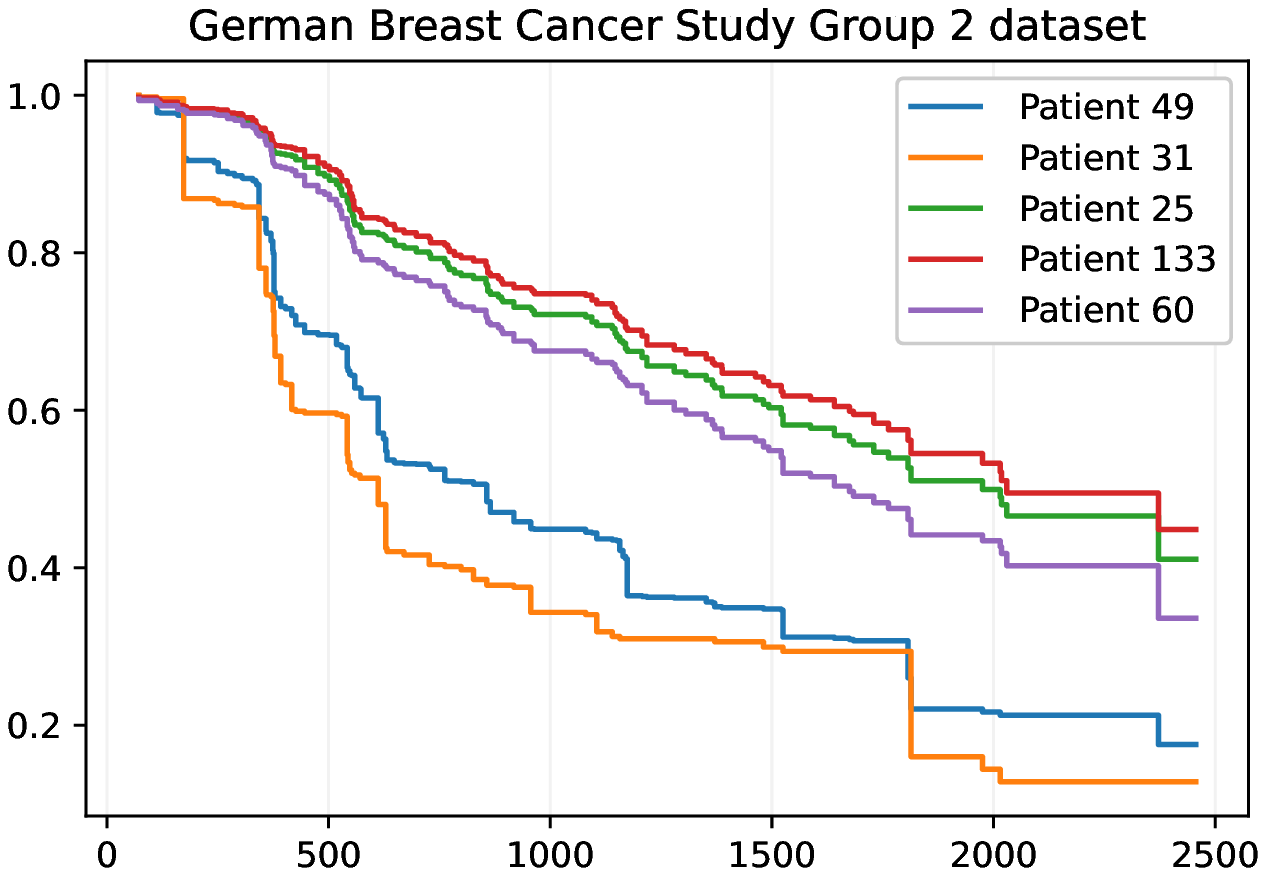}}\\
  \subfigure[$\xi_{3}$]{\includegraphics[width = 0.65\textwidth]{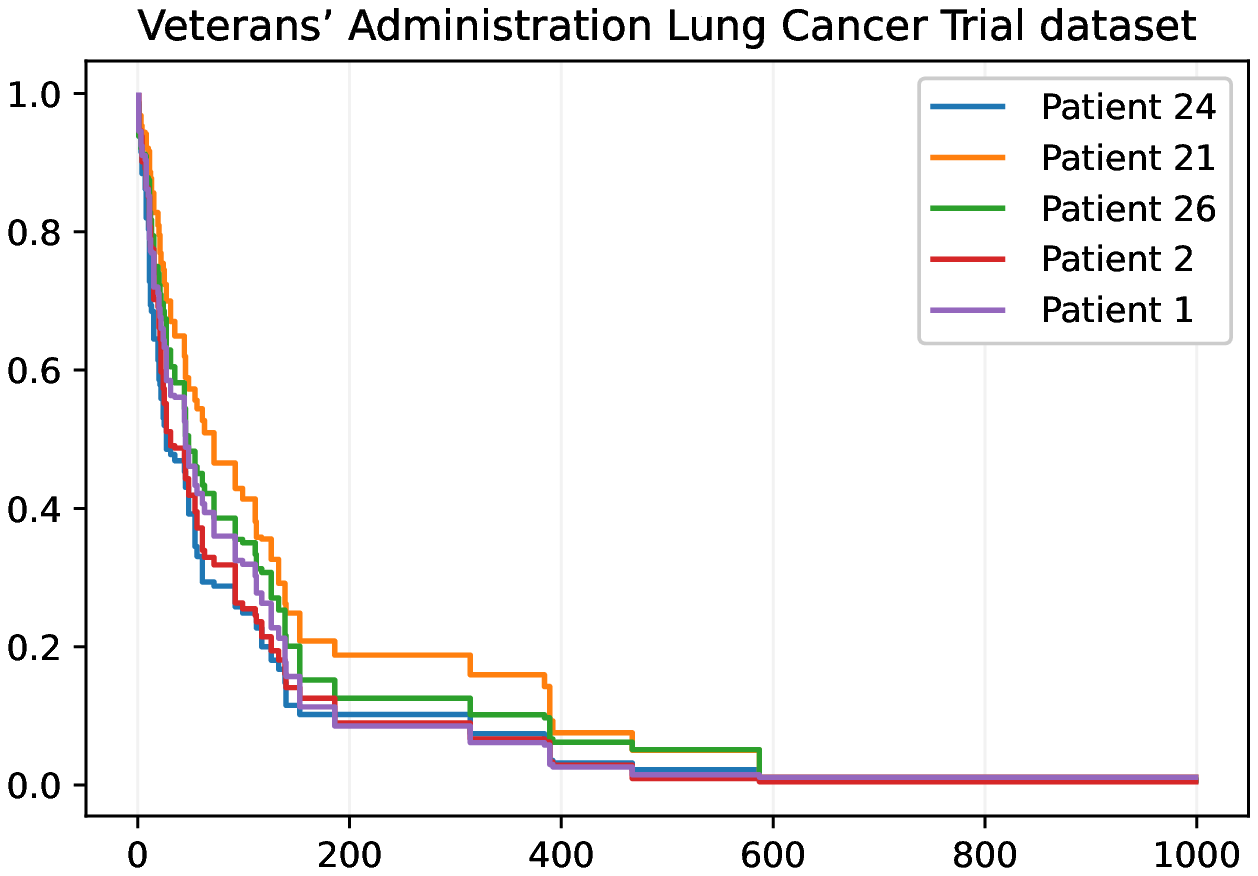}}\\
\caption{Survival Curve for Weighted Survival Cobra for three datasets with Frobenius Norm \label{fig3}}
\end{center}
\end{figure}

\begin{figure}[ht!]
 \begin{center}
  \subfigure[$\xi_{1}$]{\includegraphics[width = 0.45\textwidth]{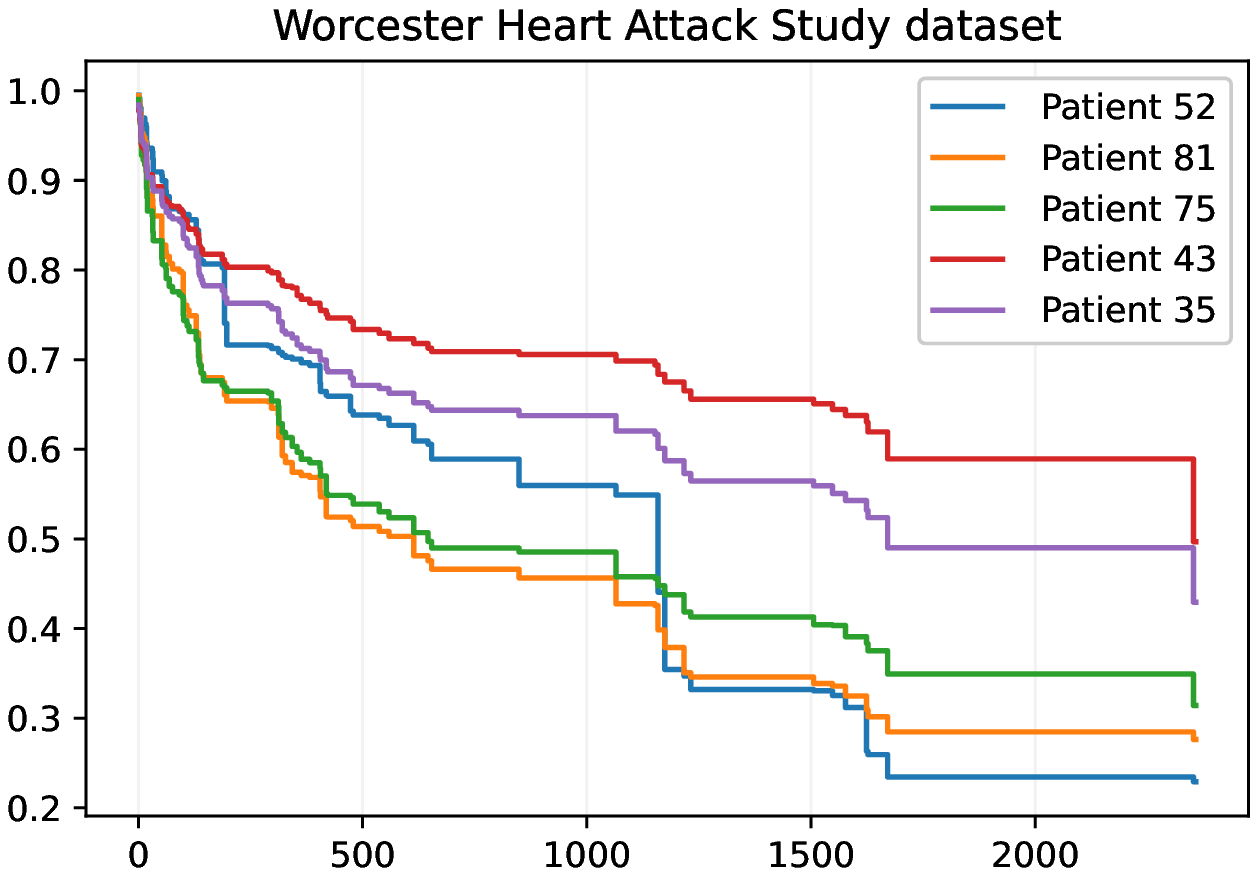}}
  \subfigure[$\xi_{2}$]{\includegraphics[width = 0.45\textwidth]{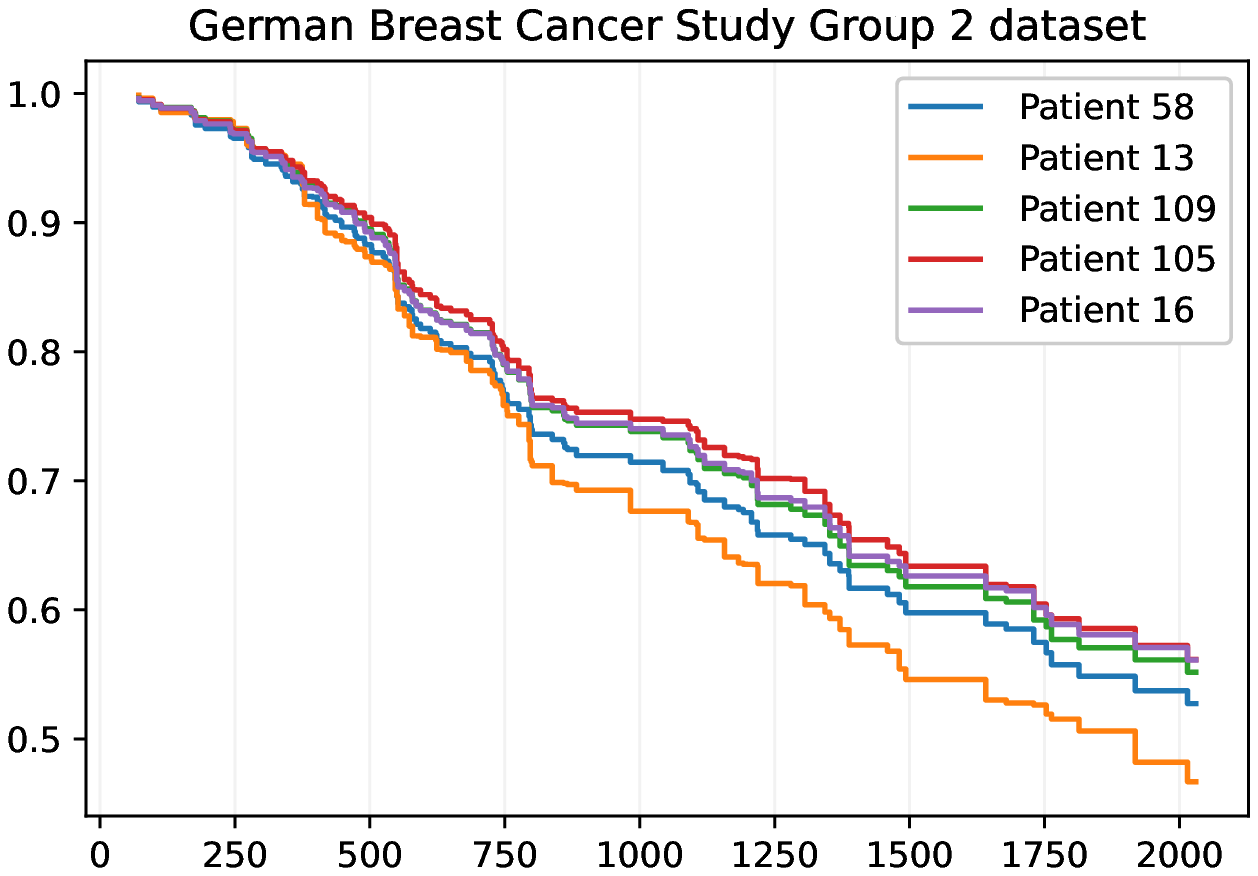}}\\
  \subfigure[$\xi_{3}$]{\includegraphics[width = 0.65\textwidth]{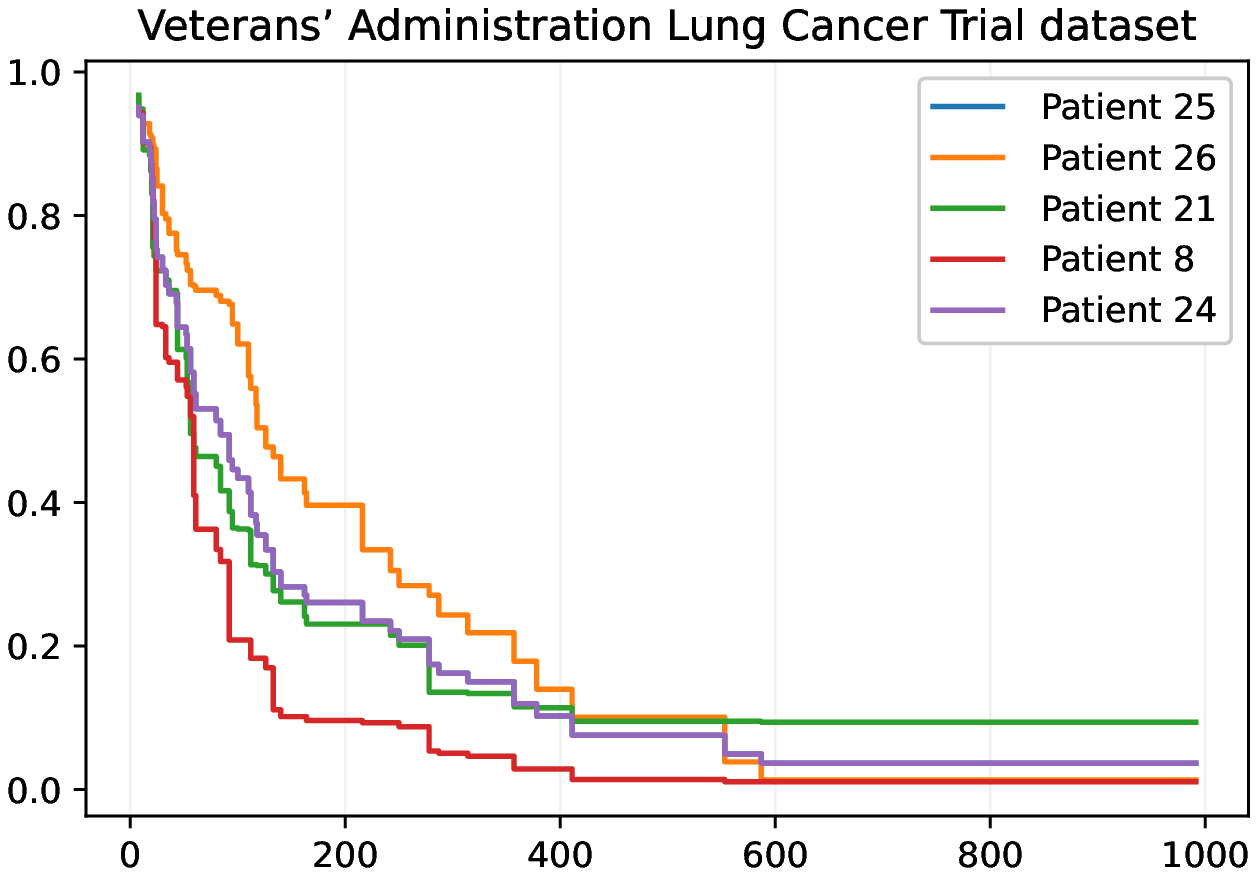}}\\
\caption{Survival Curve for Weighted Survival Cobra for three datasets with Max Norm \label{fig4}}
\end{center}
\end{figure}

  The conditional survival function obtained for different datasets are available in Figure-\ref{fig1}, Figure-\ref{fig2}, Figure-\ref{fig3} and Figure-\ref{fig4}.  The following 12 Figures show conditional survival curves for different patients in two different benchmarks and with two different COBRA implementations.  We consider eight survival trees as weak learners to construct the survival functions.  In this experiment, we choose 50\% initial division of the whole dataset to form $D_{l}$ and $D_{k}$ and try to optimize the parameters.  We choose 80\% and 20\% division ratios for the train and test set for the final prediction during the final prediction.  Two parameters take the optimal value at the cross-validated test set concordance score.  We use five-fold cross-validation to calculate the above score.

  Table-\ref{table-concor} provides loss-score calculations for different datasets and for different norms.  We denote COBRA-1 when it uses the Frobenius norm and COBRA-2 when it uses Sup-norm.  We randomly select the first consecutive weak learners and show the loss-score results.  We verify whether all score calculation matches their intuition.  On a different note, they should be higher in value than what COBRA should provide.  The calculation of mean and standard deviation of concordance loss (1 - concordance Index) presented in the table uses 1000 iterations.  Each iteration consists of a random split of 80\% and 20\% division of the train and test dataset.  We compare the results with similar ensemble techniques like Random Survival Forest and observe that all different COBRAs perform better or at the same accuracy level.  Weighted COBRA beats all algorithms not only in terms of accuracy but also due to its ability to produce survival curves.
  
\begin{table*}[ht!]
\caption{Concordance Index for different Dataset and different COBRA algorithms}
\label{table-concor}
\centering    
\begin{tabular*}{\textwidth}{@{\extracolsep{\fill}}crrrrrc} \hline \\
\toprule
& \multicolumn{2}{c}{ DATASET I} & \multicolumn{2}{c}{ DATASET II} & \multicolumn{2}{c}{ DATASET III} \\
\\\cmidrule(r){2-3}\cmidrule(lr){4-5}\cmidrule(l){6-7}
 &  mean &  & mean &  & mean &  \\
 & sd.  &       & sd.   &   & sd &  \\
\midrule
\multirow{2}{*}{Weighted-Survival COBRA 1 } & 0.2813 &  & 0.3690  & & 0.3832 & \\ 
&0.0391 && 0.040388 && 0.0634 & \\ [2ex]
\multirow{2}{*}{Weighted-Survival COBRA 2 } &  0.2705 &  & 0.350436 & & 0.3702 & \\ 
&0.039 && 0.033199 && 0.0614 & \\ [2ex]
\multirow{2}{*}{Vanilla-Survival COBRA 1 } &  0.274064	 &  & 0.370564 & & 0.423134 & \\ 
& 0.033473 && 0.039722	 && 0.078164	 & \\ [2ex]
\multirow{2}{*}{Vanilla-Survival COBRA 2 } &  0.310436 &  & 0.388382  & & 0.395362	 & \\ 
& 0.052027 && 0.044932 && 0.079831	 & \\ [2ex]
\multirow{2}{*}{Random Survival Forest } &  0.2807 &  & 0.3714  & & 0.4044	 & \\ 
& 0.0404 && 0.0405 && 0.0763	 & \\ [2ex]
\bottomrule
\end{tabular*}
\label{table:name}
\end{table*}     

\section{Conclusion}  Usual vanilla Survival CoBRA does not provide the desired result in specific norms and datasets.  However, Weighted Survival Cobra works with almost all datasets and norms.  Weighted Survival COBRA provides a better loss score too.  The information provided in this paper can be beneficial to the doctors and the insurance companies.  Since the data uses patient-specific information as covariates, real-life implementation requires data privacy protection or consent to access patient data from hospital authorities or patients, which may not allow real-life integration many times in practice.  However, we can produce the result without taking too much time.   It requires storing the training information separately before running on Mobile or other devices.  There are multiple different ways COBRA architecture can solve the same problem.  Also, the paper does not consider a cause-specific analysis in such a setup.  We explore those problems in separate research work.

\section{Declarations}  

  We keep all relevant declarations in the following subsections.  
  
\subsection*{Datasets}  All datasets are publicly available in the python module called sksurv.

\subsection*{Funding and Financial Affiliation of the work}  The work is not associated with funding agency. 

\subsection*{Conflicts of interest/Competing interests }  The content of this article contains unique contribution.  None of the author has any kind of conflict of interest in contributing for the article.  All authors have consent in contribution of the work.  

\subsection*{Availability of Code}  Authors can also provide all relevant codes on request in case of any difficulty in accessing the above repository.

\bibliographystyle{spmpsci}
\bibliography{surv}


\end{document}